\documentclass{article}
\usepackage[numbers,sort&compress]{natbib}
\usepackage{spconf}
\usepackage{amsmath,amssymb,amsfonts}
\usepackage{algorithmic}
\usepackage{graphicx}
\usepackage{textcomp}
\usepackage{xcolor}
\usepackage{booktabs}
\usepackage{stfloats}
\usepackage{subfigure} 
\usepackage[misc]{ifsym}
\usepackage{multirow}
\usepackage{array}
\usepackage{colortbl}
\usepackage{nicematrix}
\usepackage{tikz}
\usetikzlibrary{calc, tikzmark}
\usepackage{makecell}
\usepackage{nicematrix}
\usepackage{tikz}
\usetikzlibrary{calc}

\usepackage[pagebackref=true,breaklinks=true,letterpaper=true,colorlinks, citecolor=blue,bookmarks=false]{hyperref}

\title{TabMixer: Excavating Label Distribution Learning with Small-scale Features}
\name
{Weiyi Cong, Zhuoran Zheng and Xiuyi Jia$^{*}$\thanks{$*$Corresponding authors.}}
\address
{CSE, Nanjing University of Science and Technology}

\begin{document}
%
\maketitle
\begin{abstract}
Label distribution learning (LDL) differs from multi-label learning which aims at representing the polysemy of instances by transforming single-label values into descriptive degrees.
Unfortunately, the feature space of the label distribution dataset is affected by human factors and the inductive bias of the feature extractor causing uncertainty in the feature space.
Especially, for datasets with small-scale feature spaces (the feature space dimension $\approx$ the label space), the existing LDL algorithms do not perform well.
To address this issue, we seek to model the uncertainty augmentation of the feature space to alleviate the problem in LDL tasks.
Specifically, we start with augmenting each feature value in the feature vector of a sample into a vector (sampling on a Gaussian distribution function).
Which, the variance parameter of the Gaussian distribution function is learned by using a sub-network, and the mean parameter is filled by this feature value.
Then, each feature vector is augmented to a matrix which is fed into a mixer with local attention (\textit{TabMixer}) to extract the latent feature.
Finally, the latent feature is squeezed to yield an accurate label distribution via a squeezed network.
Extensive experiments verify that our proposed algorithm can be competitive compared to other LDL algorithms on several benchmarks.
\end{abstract}
\vspace{-3mm}
\section{Introduction}
\vspace{-3mm}
During the development of machine learning tasks, label distribution learning (LDL)~\cite{geng2016label} is an important machine learning paradigm that leverages a function to map a single instance to a set of labels (labels are represented in the form of descriptive degrees and the sum of the descriptive degrees is 1).
Unlike the multi-label learning paradigm, the LDL conveys a richer semantic content in terms of characterizing the instance's emotions~\cite{ZhouXG15,ZhouZZZG16} and estimating the learning task's uncertainty~\cite{zhengLABEL,le2022uncertainty,SiWPX22}.

Although several classical LDL algorithms~\cite{geng2016label,jia2018label,ren2019label,wang2021label,qian2022feature,jia2021label,RenJLCL19,ZhaoZ18a,ZhangZLX22,ChenGXZYL22,WangG19} are proposed to tackle the task of modeling the feature space into the label space, these algorithms usually favor an accurate and ample feature space.
Briefly, these algorithms expect to conduct a process of condensing the representation space rather than augmenting it.
Here, we define a lemma that a feature space of dimension $\approx$ the label space is a small-scale feature space.
A shred of evidence is that almost all the proposed LDL algorithms report weak performance on benchmark datasets with a large number of labels in many studies.
So far, we draw two questions about this: \textbf{1)} \textit{For the label space, is it difficult for the comparatively small amount of feature information to provide the algorithm with effective features to regress an accurate label distribution?} 
\textbf{2)} \textit{For the feature space, are there artificial reasons and uncertainty of the feature extractor that cause the low quality of the feature space?}
Unfortunately, we cannot parse the existing LDL dataset because the details of feature processing are blind-boxed.
Further, we want to boost the feature dimension and infer the uncertainty of the feature space by tapping into expert knowledge is costly.
To solve the above two problems, we propose a feature augmentation technique with uncertainty awareness enforced on \textit{TabMixer} (\textbf{Tab}ular MLP-\textbf{Mixer}) to learn an LDL dataset with small-scale features.
Note that our network treats tabular data equally and does not distinguish between logical and continuous values.

Overall, our approach can be grouped into the following learning cohorts.
First, to augment the feature space, an MLP-based sub-network (\textit{Learner}) is created to learn the variance of a Gaussian function.
This \textit{Learner} inputs the raw feature vectors and then assigns a unique variance value to each of the feature values in the raw feature vectors.
Combining the above, we can design a Gaussian function for each element in the raw feature space by taking the feature value as the mean value and using Gaussian sampling to obtain a vector to replace that element (the time seed is fixed in the model training phase).
By now, our input pattern is evolved from 1D to 2D and can be pseudo-considered as a grayscale map.
Subsequently, the augmented feature information is fed into \textit{TabMixer}, where each linear layer shortcut in \textit{TabMixer} is a convolution operator to capture the local characteristics of the features.
Finally, the output feature map is squeezed by the squeezed network to obtain an accurate label distribution, where the floodgates of the network use $\texttt{softmax}$.
The network is trained using the loss function of only L1 and K-L divergence.
We use two standard and a synthetic benchmark to evaluate our approach and other comparative algorithms, and the experimental results verify that the proposed algorithm is still robust under fully supervised and noisy conditions.
Furthermore, since there are random sessions in the network content, we considered a pre-training manner to eliminate this random consistency.
This paper has two key \textcolor{red}{contributions}, \textbf{i)} We propose a novel one-stop feature augmentation-learning solution executed on LDL datasets with small-scale features. \textbf{ii)} We develop a deep network (\textit{TabMixer}) that takes into account both local and global information and a new synthetic dataset.

\begin{figure*}[htbp]
	\centerline{\includegraphics[width=1.0\textwidth]{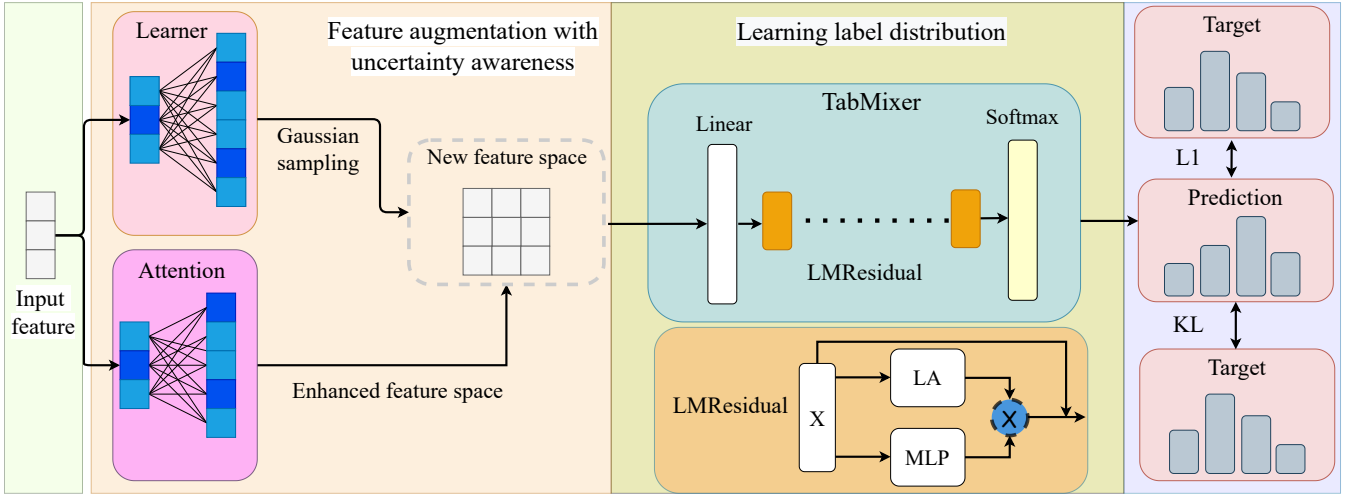}}
	\caption{\textbf{Our architecture}. Our algorithm aims to regress the label distribution of a sample using \textit{TabMixer}, where there are \textbf{two key approaches}, one is to augment the feature space by modeling uncertainty, another one is to obtain an accurate labeling distribution by mixed learning, and in addition, randomness is also considered.}
	\label{fig}
	\vspace{-4mm}
\end{figure*}

\vspace{-4mm}
\section{related work}
\vspace{-2mm}
\noindent \textbf{Label distribution learning.}
Geng et al.~\cite{geng2016label} pioneered a new machine learning paradigm: LDL, which conveys richer semantics by converting labels into descriptive degrees.
Subsequently, numerous studies~\cite{geng2016label,jia2018label,ren2019label,wang2021label,qian2022feature,jia2021label,RenJLCL19,ZhaoZ18a,ZhangZLX22,ChenGXZYL22,WangG19,zhengLABEL,le2022uncertainty,SiWPX22} are opened for LDL tasks, which involve both applications and pure theory.
One of the papers~\cite{zhengLABEL} is very interested in modeling the uncertainty of the label distribution values via deep networks.
Inspired by this, we address the feature space at small scales to model uncertainty to offer richer materials for downstream models.
%
%
%
%
%
%
%
%
%

\noindent \textbf{Tabular Learning.}
Recently there is extensive work~\cite{chen2016xgboost,ke2017lightgbm,gorishniy2021revisiting,arik2021tabnet,wang2022transtab} being proposed to model on tabular datasets.
However, these methods are usually known for the characteristics of the table's attributes.
Inspired by TransTab~\cite{wang2022transtab}, we seek to use $\texttt{MLP}$ to globally model on tabular datasets.
Furthermore, to enhance the modeling capability of the whole model, inductive bias based on convolutional operators is also fused in the network.
The architecture of the whole model is thanks to MLP-Mixer~\cite{tolstikhin2021mlp}.
%
%

\vspace{-2mm}
\section{Proposed method}
\vspace{-2mm}
The architecture of our approach is shown in Fig.~\ref{fig}.
Our approach can be described as a two-stage tactic in an end-to-end manner.
The first stage is feature augmentation with uncertainty awareness, which aims at the re-representation of the input information by embedding prior knowledge.
The purpose of the second stage is to learn the label distribution with the help of an \textit{TabMixer} in a new feature space.
Furthermore, we introduce the training strategy (loss functions) of the model and a regularization scheme (elimination of the random consistency) at the end of this section.
%
%
%
%

\noindent \textbf{Feature augmentation with uncertainty awareness.}
Given an input feature space $\mathcal{X}\in R^{m \times n}$ ($m$ is the number of instances and $n$ is the dimension of features), we assume the existence of Gaussian noise $\mathcal{N}$ in this space~\cite{mohammed2016study}.
In other words, we augment a single feature value and must consider that the source of this value may be a Gaussian distribution.
The Gaussian function has two key parameters ($\mu$ and $\sigma$), which can be formalized as:
\vspace{-4mm}
\begin{equation*}
\vspace{-1mm}
	\mathcal{N}=\frac{1}{\sqrt{2 \pi} \sigma} e^{-\frac{(x-\mu)^{2}}{2 \sigma^{2}}}, \quad x \sim \mathcal{X}.
\vspace{-1mm} 
\end{equation*}

So far, our feature augmentation method with uncertainty is based on this a priori assumption to provide more material for the downstream network.
The following describes the pipeline for this method.
For a single sample $\mathcal{V}$, we develop a \textit{Learner} to adapt a variance $\sigma_{i}$ to each element $\mathcal{V}_{i}$ in this sample.
\textit{Learner} is consist of three linear layers and three activation layers, each of which utilizes the $\texttt{ReLU}$ operator except for the last layer.
The last layer of the network layer uses $\texttt{sigmoid}$ and the dimensionality of the output layer is the same as the input layer.
Next, a sampling action is conducted where we need to construct a Gaussian distribution function $\mathcal{N}_{i}$ for each element $\mathcal{V}_{i}$.
We construct the two parameters of the Gaussian function $\mathcal{N}_{i}$ using the studied variance $\sigma_{i}$ and the feature value $\mathcal{V}_{i}$ of this sample, respectively.
Then, the execution adopts operations on these Gaussian functions, the number of sampling points is consistent with the dimensionality of the samples, and the time seed is fixed during program implementation.
%
%
%
%
%
For this newly created sample $\mathcal{G}_{i}$, it can be represented as follows:
\[\mathcal{G}_{i} = \left( \begin{array}{cccc}
	g_{1}^{1} & g_{1}^{2}&\cdots&g_{1}^{t} \\
	g_{2}^{1} & g_{2}^{2} &\cdots& g_{2}^{t} \\
	\vdots&\vdots& \ddots&\vdots \\
	g_{t}^{1} & g_{t}^{2}&\cdots & g_{t}^{t} 
\end{array}\right),\]
where each column represents a vector from the elemental values $\mathcal{V}_{i}$ of the raw samples $\mathcal{V}$.
Furthermore, to model the long-range dependence between features of the new samples, we introduce an attention mechanism acting on $\mathcal{G}_{i}$.
This attention module performs an $\texttt{MLP(sigmoid($\mathcal{G}_{i}$))}$ in the horizontal dimension of $\mathcal{G}_{i}$, where the output is multiplied by $\mathcal{G}_{i}$.
Finally, the generated feature matrix $\mathcal{G}^{*}$ is used to yield an accurate label distribution.

\noindent \textbf{Learning label distribution.}
Up to this point, an issue is raised that since the dimensionality of the feature signal is transformed from vector to matrix, a vanilla $\texttt{MLP}$ may be difficult to employ on $\mathcal{G}^{*}$.
To address this problem, we propose a mixer with local attention (\textit{TabMixer}) to extract the deep semantics of tabular data.
In a nutshell, the network treats each vector $\mathcal{G}^{*}_{i}$ as a token and relies on a mixed-learning algorithm to study the correlation between features.
To boost the modeling capability of the whole network, we develop a local attention ($\texttt{LA}$) block to be co-located with the $\texttt{MLP}$ as a component (LMResidual) in the network.
$\texttt{LA}$ consists of a 2D convolution and a $\texttt{ReLU}$, where the convolution kernel is 3, the stride size is 1, and the padding is 1.
%
%
%
So, for this module, it can be expressed as follows:
\begin{equation*}
\text{LMResidual} = \texttt{LA}(x) \times \texttt{MLP}(\texttt{LayerNorm}(x) + x),
\end{equation*}
where $x$ is the feature map and the whole operation is a structure of residuals.
%
%
Finally, $\mathcal{G}^{*}$ passes through several LMResiduals to obtain a feature map $x$. 
The feature map $x$ is fed into a squeezed network with a squeezing operator to obtain an accurate label distribution.
This squeeze network mainly consists of an $\texttt{Linear}$ and an $\texttt{Softmax}$, and the squeeze operator is a mean value function enforced on the vertical axis of the feature map $x$.

\noindent \textbf{Loss function.}
We used $L_1$ and KL divergence ($L_{KL}$) to optimize the network.
So our loss function can be expressed as follows:
$$\text{Loss} = \alpha \times L_1 + \beta \times L_{KL},$$
where $\alpha$ and $\beta$ are parameters.
In the experiment, we select $\alpha$ and $\beta$ to be 1 and 0.5, respectively.

\noindent \textbf{Eliminating random consistency.}
Foreseeably, there is an uncertainty session in our model (Gaussian sampling) that may generate pseudo-accuracy in the training stage to mislead researchers to perform early stops.
This is due to the Gaussian sampling that causes the output distribution of the model inconsistent with the distribution of the real training set.
A typical solution is to enforce a regularization term on the loss function in the training phase to remedy the accuracy calculation lemma.
In contrast, we seek to alleviate this problem using a pre-training approach.
Specifically, we remove the feature augmentation algorithm and replace it by copying several copies of the sample to concatenate it into a matrix $\mathcal{M}$ consistent with the $\mathcal{G}^{*}$ dimension.
This new sample space $\mathcal{M}$ is fed into \textit{TabMixer} to pre-train the parameters of the whole network.
Pre-training method is validated in the ablation experiment session.

\vspace{-4mm} 
\section{EXPERIMENTS}
\vspace{-3mm}

\noindent \textbf{Dataset.}
To demonstrate the capabilities of the model, we conduct experiments on three special LDL datasets (Yeast-alpha, Human Gene, and a synthetic dataset).
These datasets are all small-scale feature spaces, as shown in Table~\ref{tab:1}.
The synthetic data is based on the \textit{minist} dataset and the feature space is flattened with the help of an $\texttt{Flatten}$ operator, after which the feature space (28 $\times$ 28) is reduced to 28 using PCA.
The label space is unsampled by building a Gaussian function (mean is the class of handwriting and variance is 0.5).
When evaluating the model performance, we used the six metrics proposed by \cite{geng2016label}, which are Chebyshev distance (Chebyshev $\downarrow$), Clark distance (Clark $\downarrow$), Canberra distance (Canberra $\downarrow$), KL divergence (K-L $\downarrow$), Cosine similarity (Cosine $\uparrow$), and Intersection similarity (Intersection $\uparrow$), where $\uparrow$ represents the bigger the better, and $\downarrow$ represents the smaller the better.

\noindent \textbf{Comparative algorithms.}
We compared the results of our model with five LDL algorithms: DDH-LDL~\cite{ZhangZLX22}, BFGSLLD~\cite{geng2016label}, LDL-LRR~\cite{jia2021label}, LDLSF~\cite{RenJLCL19} and LALOT~\cite{ZhaoZ18a}.
Except for DDH-LDL, the indicators for all datasets of the models are from~\cite{jia2021label}.
Since DDH-LDL does not release the code under public resources, we attempt to replicate it using the PyTorch 1.2 platform on a single GPU shader.
\begin{table}[htbp] \footnotesize 
	\vspace{-6mm}
	\centering
	\caption{Statistics of three real-world datasets.}
		\begin{tabular}{c|ccc}
			\toprule
			Dataset & Examples & Features & Labels \\
			\midrule
			Human Gene & 17892 & 36    & 68 \\
			Yeast-alpha  & 2465   & 24   & 18 \\
			Synthetic dataset & 60000  & 28  & 56 \\
			\bottomrule
		\end{tabular}%
	\label{tab:1}%
	\vspace{-4mm}
\end{table}%
\begin{table}[htbp] \footnotesize 
	\vspace{-6mm}
	\centering
	\caption{Parameters on the three datasets.}
	\begin{tabular}{c|ccc}
		\toprule
		Dataset & Batch size & Learning rate  & Epoch  \\
		\midrule
		Human Gene & 1000   & 0.0002 & 500    \\
		Yeast-alpha  & 1000   & 0.0002 & 500    \\
		Synthetic dataset & 20000  & 0.0003 & 200   \\
		\bottomrule
	\end{tabular}%
	\label{tab:2}%
\end{table}%

\noindent \textbf{Experimental setting.}
For our algorithm, we put the customized selection of parameters for each dataset in Table~\ref{tab:2}. 
The parameters that may be used involve batch size, learning rate, and epoch. 
Our network is stacked with 12 LMResiudal on each dataset, with 512 neurons used on the $\texttt{MLP}$.
%
%
%
Besides, we use the PyTorch 1.2 framework and AdamW optimizer to train and test a deep network on the GPU shader.
For time seeds, we used $\texttt{Seed(1024)}$ on the PC (Intel(R) Xeon(R) Gold 6226R CPU @ 2.90GHz) and server (3090RTX) terminals.
Datasets are loaded using $\texttt{number\_work = 8}$ on the  PyTorch platform.

\noindent \textbf{Results and discussion.}
The experimental results for each dataset are summarized in Table~\ref{tab:3}, where we take the results of the 10 times 5-fold cross-validation. 
The experimental results are reported in the form of ``mean$\pm$std''. 
Our algorithm performs optimally on six parameters across the three datasets, particularly on the metrics (\textit{Clark}, \textit{Cosine}). 
Our approach also achieves competitive results in the remaining metrics. 
Our method has two key advantages, firstly it estimates the noise in the feature space and secondly, the deep network has robust modeling capability.
%
%
Besides, algorithms with restricted priors have better performance, such as LDL-LRR.
%
Our method has a better performance compared to LDL-LRR by taking into account the uncertainty of the feature space.
Although our modeling has an expensive cost, a GPU acceleration can be enforced in the PyTorch 1.12 to alleviate this problem.
We statistic the time \textit{TabMixer} executes inference on three datasets, including CPU and GPU.
\textit{TabMixer} is run on the Yeast-alpha dataset for \{0.04s, 0.02s\}, on the Human Gene dataset for \{0.06s, 0.02s\}, and on the Synthetic dataset for \{0.008s, 0.01s\} on the CPU and GPU, respectively.
%
%

\begin{table*}[htbp] \footnotesize 
	\centering
	\caption{Experimental results on three datasets and the best results are bolded.}
	\resizebox{\linewidth}{!}{
		\begin{tabular}{c|c|cccccc}
			\toprule
			\textbf{Dataset} & \textbf{Algorithm} & \multicolumn{1}{c|}{\textbf{Chebyshev $\downarrow$}} & \multicolumn{1}{c|}{\textbf{Clark $\downarrow$}} & \multicolumn{1}{c|}{\textbf{Canberra $\downarrow$}} & \multicolumn{1}{c|}{\textbf{K-L $\downarrow$}} & \multicolumn{1}{c|}{\textbf{Cosine $\uparrow$}} & \textbf{Intersection $\uparrow$} \\
			\midrule
			\multirow{6}[2]{*}{Synthetic dataset} & Ours  & \textbf{0.1133$\pm$0.0065} & \textbf{0.5012$\pm$0.0014} & \textbf{0.9787$\pm$0.0040} & 0.1061$\pm$0.0021 & \textbf{0.9596$\pm$0.0072} & \textbf{0.8491$\pm$0.0066} \\
			& DDH-LDL & 0.1541$\pm$0.0621 & 0.6431$\pm$0.2224 & 1.2002$\pm$0.0034 & 0.1451$\pm$0.0122 & 0.9039$\pm$0.0044 & 0.7953$\pm$0.0923 \\
			& LDL-LRR & 0.1150$\pm$0.0013 & 0.5231$\pm$0.0065 & 0.9996$\pm$0.0093 & \textbf{0.0982$\pm$.0031} & 0.9153$\pm$0.0020 & 0.8355$\pm$0.0023 \\
			& BFGS-LLD & 0.1396$\pm$0.0225 & 0.5856$\pm$0.0494 & 1.1309$\pm$0.1098 & 0.1372$\pm$0.0374 & 0.9108$\pm$.0227 & 0.8153$\pm$0.0263 \\
			& LALOT & 0.2243$\pm$0.0912 & 1.334$\pm$0.800 & 2.9023$\pm$0.8996 & 3.5516$\pm$0.9542 & 0.7020$\pm$0.1415 & 0.7015$\pm$0.2214 \\
			& LDLSF & 0.1270$\pm$0.0011 & 0.6227$\pm$0.0112 & 1.0146$\pm$0.0158 & 0.1936$\pm$0.0257 & 0.9220$\pm$0.0028 & 0.8280$\pm$0.0034 \\
			\midrule
			\multirow{6}[2]{*}{Yeast-alpha} & Ours  & \textbf{0.0113$\pm$0.0025} & \textbf{0.2044$\pm$0.0071} & \textbf{0.6601$\pm$0.1522} & 0.0061$\pm$0.0112 & \textbf{0.9948$\pm$0.0087} & \textbf{0.9633$\pm$0.0052} \\
			& DDH-LDL & 0.1627$\pm$0.0529 & 0.5228$\pm$0.1937 & 1.0236$\pm$0.3634 & 0.1177$\pm$0.0638 & 0.8794$\pm$0.0936 & 0.8128$\pm$0.9355 \\
			& LDL-LRR & 0.0134$\pm$0.0002 & 0.2093$\pm$0.0031 & 0.6791$\pm$0.0100 & \textbf{0.0054$\pm$0.0001} & 0.9946$\pm$0.0001 & 0.9625$\pm$0.0005 \\
			& BFGS-LLD & 0.0135$\pm$0.0001 & 0.2110$\pm$0.0030 & 0.6865$\pm$0.0106 & 0.0055$\pm$0.0001 & 0.9949$\pm$0.0001 & 0.9629$\pm$0.0006 \\
			& LALOT & 0.0165$\pm$0.0003 & 0.2608$\pm$0.0043 & 0.8544$\pm$0.0150 & 0.0084$\pm$0.0003 & 0.9917$\pm$0.0003 & 0.9526$\pm$0.0008 \\
			& LDLSF & 0.0139$\pm$.0002 & 0.2164$\pm$0.0021 & 0.6874$\pm$0.0111 & 0.0058$\pm$0.0001 & 0.9943$\pm$0.0001 & 0.9613$\pm$.00004 \\
			\midrule
			\multirow{6}[2]{*}{Human Gene} & Ours  & \textbf{0.0512$\pm$0.0066} & \textbf{2.0100$\pm$1.1254} & \textbf{13.4451$\pm$0.1152} & \textbf{0.2232$\pm$0.0347} & \textbf{0.8452$\pm$0.0081} & \textbf{0.7949$\pm$0.0088} \\
			& DDH-LDL & 0.0597$\pm$0.0752 & 2.8830$\pm$0.8514 & 19.9391$\pm$0.9627 & 0.3815$\pm$0.3870 & 0.7313$\pm$0.8355 & 0.6878$\pm$0.8845 \\
			& LDL-LRR & 0.0532$\pm$0.0011 & 2.1114$\pm$0.0122 & 13.5681$\pm$0.0025 & 0.2365$\pm$0.0049 & 0.8346$\pm$0.0020 & 0.7844$\pm$0.0014 \\
			& BFGS-LLD & 0.0539$\pm$0.0009 & 2.1270$\pm$0.0141 & 14.5633$\pm$0.1107 & 0.2398$\pm$0.0038 & 0.8328$\pm$0.0018 & 0.7828$\pm$0.0014 \\
			& LALOT & 0.0573$\pm$0.0012 & 4.0703$\pm$3.6917 & 17.8198$\pm$3.7167 & 0.2956$\pm$0.0059 & 0.8055$\pm$0.0024 & 0.7578$\pm$0.0016 \\
			& LDLSF & 0.0533$\pm$0.0009 & 2.1295$\pm$0.0209 & 14.5681$\pm$0.0055 & 0.2395$\pm$0.0054 & 0.8332$\pm$0.0028 & 0.7828$\pm$0.0022 \\
			
			\bottomrule
		\end{tabular}%
	}
	\label{tab:3}%
	\vspace{-4mm}
\end{table*}%

\noindent \textbf{Ablation studies.}
We conduct ablation studies to demonstrate the effectiveness of feature augmentation and pre-training algorithms on dataset \textit{Human Gene}.  
The results of simple ablation experiments are summarized in Table~\ref{tab:4}. 
w/o FA is an algorithm for removing the feature augmentation (the replacement scheme is a reference to the augmentation method of the data in the elimination random consistency section) and w/o PT is defined when the pre-training algorithm is removed. 
The demise of the data illustrates the effectiveness of the feature augmentation and the pre-training algorithms. 
We conduct 10 times 5-fold cross-validation on the dataset of the ablation studies.
The \textcolor{gray}{gray} markers represent optimal performance.

\noindent \textbf{Noise interference.}
To verify the robustness of our algorithm against the noise and uncertainty in feature space, we experiment on the dataset~\textit{Human Gene} with Gaussian noise.
We synthesize a dataset on~\textit{Human Gene} by Gaussian function to enforce different degrees of noise.
%
For feature space on the dataset, the variance of the Gaussian distribution function is selected from (0.1, 0.5, and, 1), and the mean value uses 0. 
The results of simple anti-noise experiments are summarized in Table~\ref{tab:5}. 
Our method still has good competition up to a noise level of 1, while other algorithms perform noise disturbances with performance degradation far beyond our method.

\begin{table}[htbp] \footnotesize 
	\vspace{-6mm}
	\centering
	\caption{Results on ablation studies.}
		\begin{tabular}{c|ccc}
			\toprule
			\textbf{Measures
			} & \multicolumn{1}{c|}{Ours} & \multicolumn{1}{c|}{w/o FA} & w/o  PT \\
			\midrule
			Chebyshev $\downarrow$ & \cellcolor{lightgray}0.0512$\pm$0.0066 & 0.0536$\pm$0.0012 & 0.5249$\pm$0.0130 \\
			Clark $\downarrow$ & \cellcolor{lightgray}2.0100$\pm$1.2554 & 2.1100$\pm$1.1010 & 2.0571$\pm$0.0066 \\
			Canberra $\downarrow$ & \cellcolor{lightgray}13.4451$\pm$0.1152 & 14.1222$\pm$0.1466 & 13.4989$\pm$0.0023 \\
			K-L $\downarrow$ & \cellcolor{lightgray}0.2232$\pm$0.0347 & 0.2281$\pm$0.0051 & 0.2293$\pm$0.0020 \\
			Cosine $\uparrow$ & \cellcolor{lightgray}0.8452$\pm$0.0081 & 0.8440$\pm$0.0070 & 0.8227$\pm$0.0845 \\
			Intersection $\uparrow$ & \cellcolor{lightgray}0.7949$\pm$0.0088 & 0.7799$\pm$0.0088 & 0.7890$\pm$0.0014 \\
			\bottomrule
		\end{tabular}%

	\label{tab:4}%
\end{table}%

\begin{table}[htbp] \footnotesize 
	\vspace{-8mm}
	\centering
	\caption{Results on noise interference.}

		\begin{tabular}{c|ccc}
			\toprule
			\textbf{Measures
			} & \multicolumn{1}{c|}{0.1} & \multicolumn{1}{c|}{0.5} & \multicolumn{1}{c}{1.0}\\
			\midrule
			Chebyshev $\downarrow$  & 0.0521$\pm$0.0043  & 0.0556$\pm$0.0011  & 0.6149$\pm$0.0100 \\
			Clark $\downarrow$      & 2.0400$\pm$1.1133  & 2.1945$\pm$1.1246  & 2.2471$\pm$0.0014 \\
			Canberra $\downarrow$   & 13.4661$\pm$0.1217 & 14.3111$\pm$0.1123 & 14.9989$\pm$0.0099 \\
			K-L $\downarrow$        & 0.2431$\pm$0.0202  & 0.2599$\pm$0.0053  & 0.2943$\pm$0.0033 \\
			Cosine $\uparrow$       & 0.8369$\pm$0.0086  & 0.8179$\pm$0.0070  & 0.7227$\pm$0.0444 \\
			Intersection $\uparrow$ & 0.7712$\pm$0.0088  & 0.7411$\pm$0.0071  & 0.6109$\pm$0.0055 \\
			\bottomrule
		\end{tabular}%
	\label{tab:5}%
	\vspace{-4mm}
\end{table}%

\vspace{-2mm}
\section{CONCLUSION}
\vspace{-2mm}
How to extract enough information from fewer features is a challenge in LDL.
In this paper, we propose an end-to-end LDL method to address it.
Our method is performed with the help of \textit{TabMixter} augmentation considering the feature space uncertainty.
The experimental results show that our model achieves good results on all six metrics.
Moreover, our method is further boosted by taking into account the random consistency factor of the model.
%
%
Finally, we evaluate the noise tolerance of the model on the benchmark dataset.
%
\vfill\pagebreak

 
\bibliographystyle{IEEEbib}
\bibliography{Template}

\end{document}